\documentclass[10pt,twocolumn,letterpaper]{article}

\usepackage{wacv}              

\usepackage[accsupp]{axessibility}  

\usepackage{graphicx}
\usepackage{amsmath}
\usepackage{amssymb}
\usepackage{booktabs}
\usepackage{array,multirow}
\usepackage{gensymb}
\usepackage[usenames, dvipsnames]{color}
\usepackage{xcolor}
\usepackage{tikz}

\DeclareRobustCommand{\tikzxmark}{%
\tikz[scale=0.23] {
    \draw[line width=0.7,line cap=round] (0,0) to [bend left=6] (1,1);
    \draw[line width=0.7,line cap=round] (0.2,0.95) to [bend right=3] (0.8,0.05);
}}

\usepackage[pagebackref,breaklinks,colorlinks]{hyperref}

\usepackage[capitalize]{cleveref}
\crefname{section}{Sec.}{Secs.}
\Crefname{section}{Section}{Sections}
\Crefname{table}{Table}{Tables}
\crefname{table}{Tab.}{Tabs.}

\def\wacvPaperID{1760} 
\def\confName{WACV}
\def\confYear{2024}

\begin{document}

\title{Toward Planet-Wide Traffic Camera Calibration}
\author{Khiem Vuong \quad \quad Robert Tamburo \quad \quad Srinivasa G. Narasimhan\\
\tt \small \{kvuong,rtamburo,srinivas\}@andrew.cmu.edu\\
Carnegie Mellon University
}

\twocolumn[{%
\renewcommand\twocolumn[1][]{#1}%
\maketitle
\vspace{-1cm}
\begin{center}
    \centering
    \captionsetup{type=figure}
    \includegraphics[width=\textwidth]{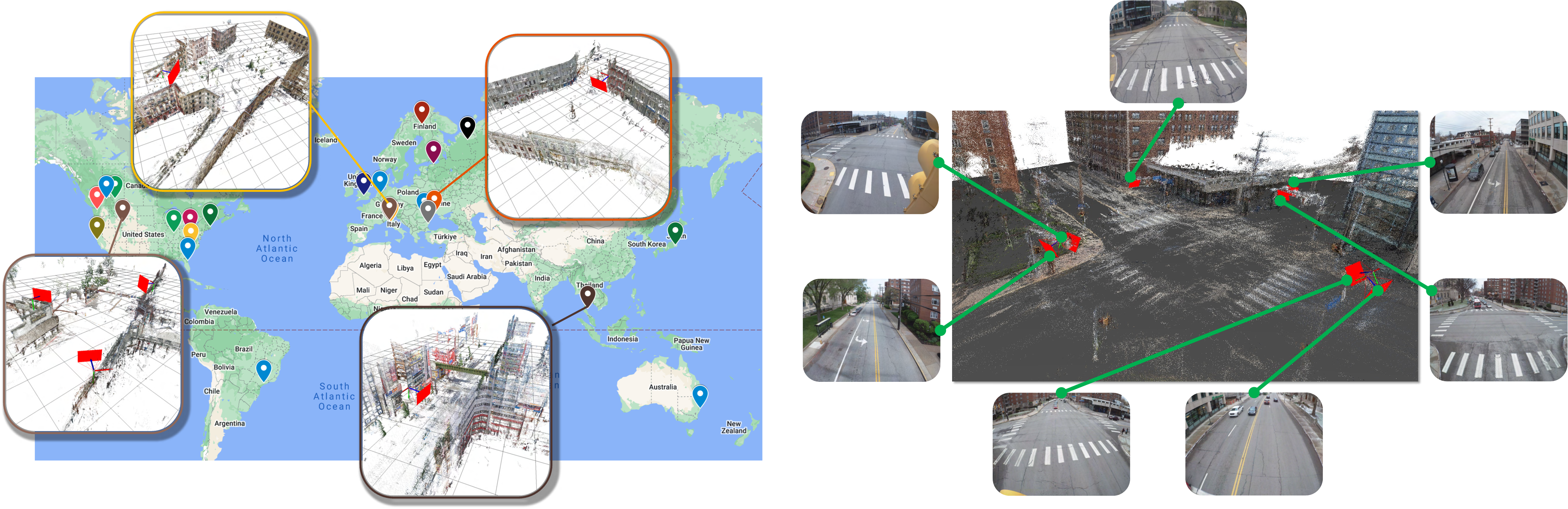}
    \captionof{figure}{Our framework allows us to perform 3D scene reconstruction and precise localization of over 100 real-world traffic cameras distributed globally across multiple countries, with the potential to scale to any camera with sufficient street-level imagery. \textbf{(Left)}: Highlighting the reconstruction and localization of traffic cameras at specific chosen locations. \textbf{(Right)}: Demonstrating 7 cameras positioned within an urban intersection, accurately localized with respect to the reconstructed 3D scene. \textbf{Please zoom in for better visualization.}}
\end{center}%
}]


\begin{abstract}
   Despite the widespread deployment of outdoor cameras, their potential for automated analysis remains largely untapped due, in part, to calibration challenges. The absence of precise camera calibration data, including intrinsic and extrinsic parameters, hinders accurate real-world distance measurements from captured videos. To address this, we present a scalable framework that utilizes street-level imagery to reconstruct a metric 3D model, facilitating precise calibration of in-the-wild traffic cameras. Notably, our framework achieves 3D scene reconstruction and accurate localization of over 100 global traffic cameras and is scalable to any camera with sufficient street-level imagery. For evaluation, we introduce a dataset of 20 fully calibrated traffic cameras, demonstrating our method's significant enhancements over existing automatic calibration techniques. Furthermore, we highlight our approach's utility in traffic analysis by extracting insights via 3D vehicle reconstruction and speed measurement, thereby opening up the potential of using outdoor cameras for automated analysis. Code and dataset will be available on the \href{https://www.khiemvuong.com/OpenTrafficCam3D}{project website}.
   
   
\end{abstract}

\section{Introduction}
\label{sec:introduction}

With the recent advances in vision techniques, traffic cameras have gained numerous applications, including vehicle speed measurement~\cite{giannakeris2018speed, sochor2018comprehensive}, automated traffic analytics~\cite{sabbani2018deep, nubert2018traffic}, and accident/anomalies detection~\cite{giannakeris2018speed, bi2019joint, shah2018cadp}, just to name a few. 
To enable such applications, camera calibration is a crucial requirement.
Camera calibration includes estimating both intrinsic parameters (focal length and distortion coefficients) and extrinsic parameters (orientation and position of the camera) in metric real-world coordinates. In addition, for many downstream applications, the metric geometry of the scene, including the ground plane is often necessary. 
However, such calibration information is not readily available in most cases. 

Despite extensive literature on traffic camera calibration, existing approaches suffer from various limitations. Traditional methods, such as those relying on checkerboard-based calibration~\cite{zhang2000flexible, tsai1987versatile}, are not practically scalable since physical access to the scene is required. Other techniques require manual inputs, like identifying landmarks with known dimensions like road markings which can be time-consuming and subject to human error~\cite{cathey2005, wang2007research}. Certain approaches rely on estimating or assuming specific priors, such as vanishing points~\cite{dubska2014, sochor2017traffic, kocur2021traffic}, average vehicle size~\cite{dailey2000}, or camera height~\cite{wang2007research}, and can introduce inaccuracies with limited generalizability to novel scenarios.

To address these challenges, we introduce a novel framework for automatically acquiring accurate metric 3D scene reconstruction and calibration of stationary traffic cameras at real-world street intersections. To achieve this, we leverage the vast amount of high-quality, geo-referenced, and calibrated images available in Google Street View (GSV)~\cite{GoogleStreetView}. By utilizing GSV, we construct a metric-scale 3D scene reconstruction near the desired traffic camera location.
Given that GSV offers panorama images, we improve our 3D reconstruction by enforcing known relative poses among perspective images sampled from the same panorama.
Next, we employ state-of-the-art (SOTA) camera localization techniques, leveraging recent advances in learned feature matching, such as SuperPoint~\cite{detone2018superpoint} and SuperGlue~\cite{sarlin2020superglue}, to establish robust 2D-3D correspondences. This enables us to infer the traffic camera's intrinsic and extrinsic parameters accurately. 
It is worth noting that while our emphasis in this paper is on the utilization of GSV, our framework is adaptable to any source of street-level imagery, including Bing's Streetside~\cite{BingStreetside}, Mapillary Maps~\cite{Mapillary}, user-captured data from smartphones (with GPS information), and other similar sources. This aspect highlights the scalability and versatility of our method, further extending its potential to achieve planet-wide camera calibration.

We demonstrate 3D scene reconstruction and accurate localization at over 100 traffic cameras across multiple countries and continents, with the potential to generalize to any novel camera where sufficient nearby street-level imagery is available.
For quantitative validation, we propose a new dataset containing 20 fully calibrated traffic cameras at diverse urban scenes under varying capture conditions.
Through extensive quantitative and qualitative experiments, we demonstrate the significant improvements of our method over existing SOTA methods in both intrinsic and extrinsic calibration.
Leveraging accurate calibration, we illustrate its capabilities in the domains of 3D reconstruction and velocity measurement for moving vehicles, thus facilitating the extraction of valuable insights from the data.

To summarize, our main contributions are:
\begin{itemize}
    \item We develop a scalable framework utilizing street-level imagery to create precise 3D models for accurate global calibration of traffic cameras, with successful localization of over 100 cameras and potential for broader application.
    \item We introduce a novel dataset featuring 20 fully calibrated traffic cameras, capturing diverse urban scenes under varying conditions, serving as a valuable benchmark for future research.
    \item We demonstrate the framework's efficacy in traffic analysis through automated extraction of insights via 3D vehicle reconstruction and speed measurement.
\end{itemize}
%


\section{Related Work}
\label{sec:related_work}

Camera calibration involves two key aspects: 1) \textit{intrinsics calibration}, which takes into account perspective projection (focal length, principal points, etc.) and potentially corrects for radial and tangential distortion, and 2) \textit{extrinsics calibration}, which refers to camera rotation and translation, usually defined with respect to the ground plane. Note that for practical real-world applications like speed measurement, the extrinsic parameters must be expressed in \textit{metric} units, often referred to as the \textit{metric scene scale}.


\noindent \textbf{Generic Camera Calibration:} Within the domain of camera calibration, several methods have been established. Two widely employed gold-standard techniques, presented by Zhang et al.\cite{zhang2000flexible} and Tsai et al.\cite{tsai1987versatile}, utilize planar calibration targets to estimate intrinsic and extrinsic camera parameters. Although those methods achieve sub-pixel calibration accuracy, they prove infeasible for traffic cameras positioned in challenging and potentially inaccessible locations, thereby limiting their scalability and practicality. Additionally, despite the capability of learning-based methods~\cite{bogdan2018deepcalib, lopez2019deep} to recover focal length and distortion parameters from a single image, they usually do not generalize well to out-of-distribution data such as traffic cameras.

\begin{figure*}[ht]
    \centering
    \includegraphics[width=\textwidth]{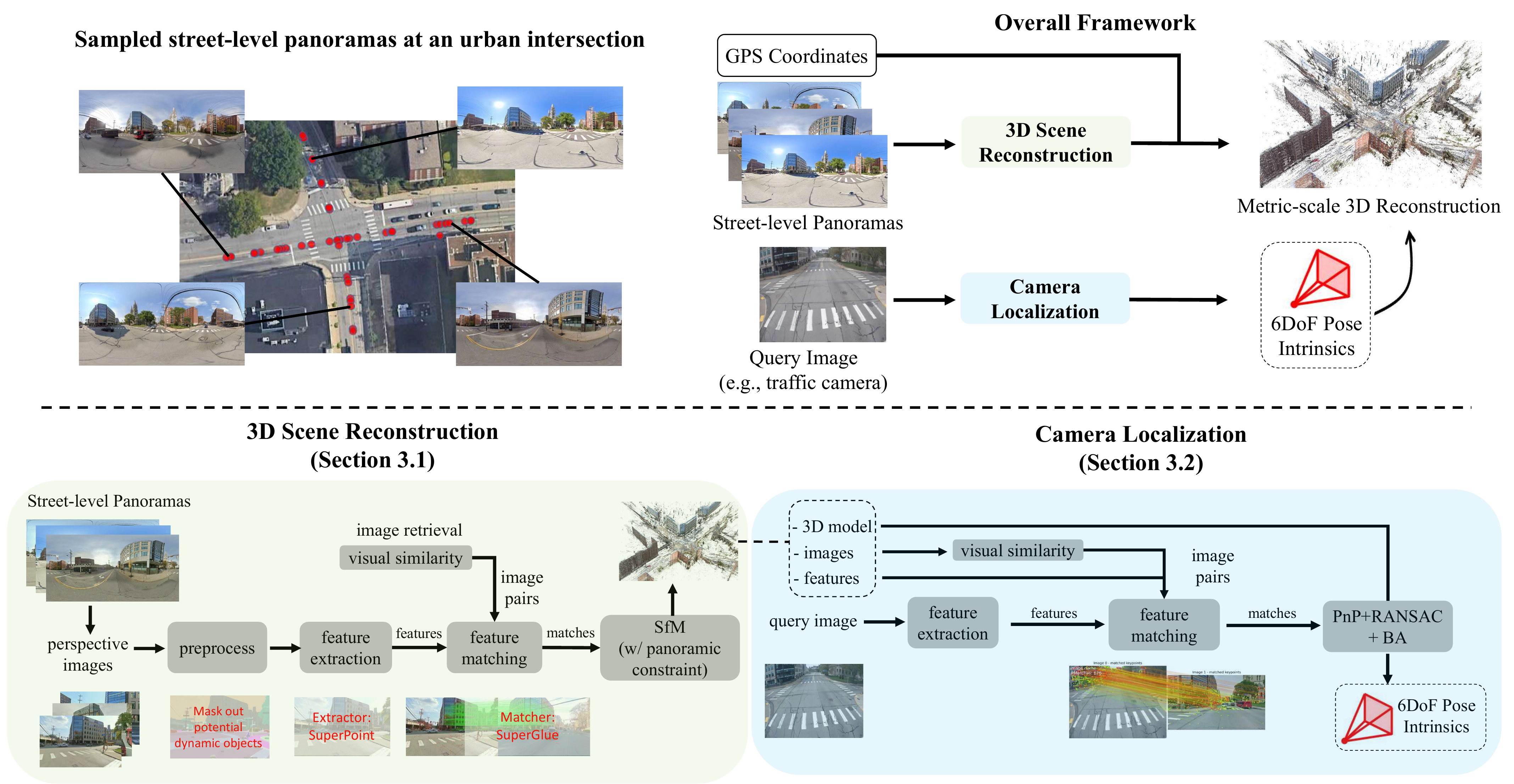}
    \caption{\textbf{Top}: Using street-level panoramas and GPS data from GSV, we reconstruct the scene in 3D for a metric-scale representation. With a query image from a traffic camera, we perform camera localization to determine intrinsic parameters and camera pose w.r.t. the 3D scene. \textbf{Bottom}: More details on \textit{3D Scene Reconstruction} (left) and \textit{Camera Localization} (right).} 
    \label{fig:calib_pipeline}
\end{figure*}

\noindent \textbf{Traffic Camera Calibration:} In the context of traffic scene analysis, a review of available methods has been presented by Sochor et al.~\cite{sochor2018comprehensive}. Some approaches~\cite{he2007b, cathey2005, grammatikopoulos2005} rely on detecting vanishing points at road marking intersections, utilizing vehicle motion to calibrate the camera~\cite{dubska2014,dubska2015,schoepflin2003,dailey2000}, or involving manual measurements of dimensions on the road plane~\cite{maduro2008,nurhadiyatna2013,sina2013,luvizon2014,luvizon2016,do2015,lan2014}. Various techniques have also been proposed for estimating the scene scale. For example, \cite{dubska2014} employed a 3D bounding box around vehicles and their average dimensions to compute the scale, and \cite{sochor2017traffic} suggested using the alignment of a 3D model and a bounding box for scale inference.
However, it is important to note that these methods are not without limitations, particularly in terms of scalability and accuracy. Manual techniques demand labor-intensive measurements of landmarks and dimensions. Meanwhile, automatic approaches relying on vehicle 3D model or vanishing points~\cite{bartl2020planecalib, kocur2021traffic, dubska2015, sochor2017traffic, bhardwaj2018autocalib} still manifest notable errors and sensitivity to the quality of estimated geometric cues, especially when certain assumptions are compromised, e.g., non-straight vehicle motion, pronounced camera distortion, different viewpoints, or lack of knowledge of the exact make/model of vehicles, etc. 
On the other hand, our method does not make any assumptions about scene geometry or vehicle motion and instead takes advantage of the extensive collection of geo-registered panoramic street-level imagery, offering a novel, practical, and scalable solution for accurate camera calibration.

\noindent \textbf{Traffic Camera Applications:} A comprehensive survey on \textit{Monocular Visual Traffic Surveillance} has been conducted by Zhang et al.~\cite{zhang2022monocular}. While applications like vehicle counting rely on 2D data, tasks demanding 3D insights such as speed estimation and distance measurement rely on precise camera calibration. In the context of speed measurement, recent methods\cite{bartl2020planecalib, revaud2021robust, kocur2021traffic} excel within specific scenarios but face limitations in generalizing to unfamiliar data. Our approach provides a simple yet effective solution, automating the acquisition of camera intrinsics, extrinsics, and metric scene geometry. This fosters the integration of 3D techniques into automated traffic analysis.


\section{Method}
\label{sec:method}


Our first objective is to construct a metric 3D reconstruction of the scene surrounding a chosen traffic camera's location, typically an intersection (Section~\ref{subsection:scene_reconstruction}). Following this, our goal is to localize the traffic camera within the reconstructed environment, thereby extracting both the intrinsic and extrinsic parameters of the camera (Section~\ref{subsection:localization}). The overall framework of our approach is depicted in Figure~\ref{fig:calib_pipeline}.

\subsection{3D Scene Reconstruction}
\label{subsection:scene_reconstruction}



To perform the reconstruction, we leverage Google Street View (GSV)~\cite{GoogleStreetView} to build the scene's geometry around a specific GPS location. GSV is a street-level imagery database and a rich source of millions of panorama images with wide coverage all over the world (more than 10 million miles across 100 countries~\cite{GSVAnniversary}). Every panorama image is geo-tagged with accurate GPS coordinates, capturing 360\degree~horizontal and 180\degree~vertical field-of-view (FoV) with high resolution (see top left of Figure~\ref{fig:calib_pipeline}). An overview of our \textbf{3D Scene Reconstruction} pipeline is illustrated in bottom left of Figure~\ref{fig:calib_pipeline}.
\begin{figure*}[ht]
    \centering
    \includegraphics[width=0.90\textwidth]{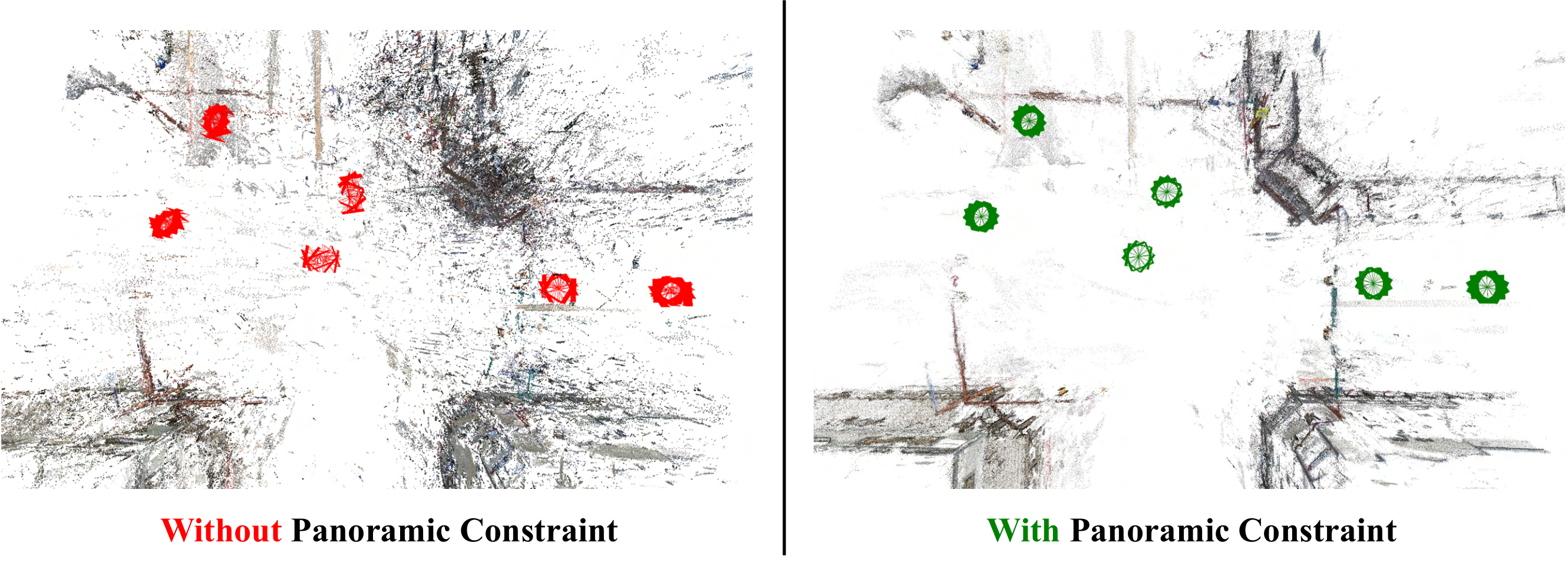}
    \vspace{-0.3cm}
    \caption{Enforcing known relative poses between perspective images from the same panorama leads to more accurate 3D reconstruction, especially in sparse-view scenarios. The recovered viewpoints in \textcolor{red}{red} (left) belonging to the same panorama do not coincide. \textbf{Please zoom in for better visualization.}}
    \label{fig:ba_constraint}
\end{figure*}

In particular, we first sample $N$ panoramas (equirectangular frames) $\mathcal{E} = \{\mathcal{E}_i | i = 1 \dots N\}$ around the desired camera's location inside a radius of 40 meters. 
Since most components of a typical \textit{structure-from-motion} (SfM) pipline~\cite{schoenberger2016sfm} are primarily optimized for perspective images, we extract ideal, pinhole camera-style perspective projections from equirectangular images before performing 3D reconstruction. 
Specifically, from each equirectangular image $\mathcal{E}_i$, we extract $T$ perspective images $\mathcal{I} = \{\mathcal{I}_{ij} | i = 1 \dots N, j = 1 \dots T\}$ that are uniformly sampled along the yaw direction with specified size and FoV, covering 360\degree~horizontal FoV. Denoting $\mathbf{\Pi}\left(\cdot\right)$ as the projection function from equirectangular to perspective image, we can define each perspective image $\mathcal{I}_{ij}$ as: 


\begin{align}
    \mathcal{I}_{ij} = \mathbf{\Pi}\Big( \mathcal{E}_i, \texttt{pitch=0}, \texttt{yaw=}\frac{2\pi * j}{T}, \texttt{fov=FOV}\Big)
\label{eq:perspective_projection}
\end{align}

In practice, with image size of $(1920\times 1080)$, we found the set of hyperparameters $\{\texttt{T=}12, \texttt{FOV=}90\degree\}$ to produce high-quality perspective images with sufficient overlap and minimal perspective distortions.
Subsequently, we adapt COLMAP~\cite{schoenberger2016sfm} to estimate camera pose $(R_{i,j}, t_{i,j})$ for each frame $\mathcal{I}_{ij}$.

\textbf{Preprocessing.} As dynamic objects often cause errors in the reconstruction, we apply a semantic segmentation method~\cite{cheng2021mask2former} to segment out potential dynamic objects such as vehicles and people and suppress feature extraction in these areas. 
For each perspective image (extracted from panorama image), the intrinsic camera matrix $K_{ij}$ is known. Therefore, we fixed the shared camera intrinsics for all the frames during reconstruction.

\textbf{Feature extraction and matching.} We adopt  SuperPoint~\cite{detone2018superpoint} and SuperGlue~\cite{sarlin2020superglue} to establish correspondences among feature points across images. Instead of using exhaustive matching where each image is matched against every other image, we employ an adapted version of vocabulary tree matching, wherein each image is matched against its nearest visual neighbors through a vocabulary tree. To build the vocabulary tree, we first compute the descriptor centroids using \texttt{KMeans++}~\cite{10.5555/1283383.1283494}, then \texttt{KDTree}~\cite{friedman1977algorithm} is used to build the vocabulary tree using VLAD~\cite{arandjelovic2013all} descriptors. The vocabulary tree serves as a visual database enabling retrieval of database images that closely resemble the query image in terms of visual appearance.

\textbf{Enforcing panoramic constraints for bundle adjustment.}  In the conventional SfM workflow, unordered input image collections lead to the independent treatment of each image. However, in our scenario where perspective images stem from panorama sampling, we can leverage the known transformations or relative poses between frames that are sampled from the same panorama.
To capitalize on this, we augment the typical Bundle Adjustment~\cite{triggs2000bundle} (BA) by incorporating the known relative poses between frames within the same panorama. In our context, two perspective images from a common panorama are linked by a pure rotation around the $z$-axis (i.e., along the yaw direction, refer to Eq.~\ref{eq:perspective_projection}). In particular, for a perspective image $\mathcal{I}_{ij}$ characterized by its extrinsic camera parameters $(R_{i,j}, t_{i,j})$, we introduce an additional optimization objective $\mathcal{L}_{\mathrm{pano}} = \mathcal{L}_{\mathrm{trans}} + \mathcal{L}_{\mathrm{rot}}$, where:
\begin{align}
\begin{split}
    & \mathcal{L}_{\mathrm{trans}} = \sum_{i=1}^{N} \sum_{j=2}^{T} ||t_{i,j} - t_{i, j-1}||^2, \\
    & \mathcal{L}_{\mathrm{rot}} = \sum_{i=1}^{N} \sum_{j=2}^{T} ||R_{i,j}^\top R_{i,j - 1} - R_z\left(\frac{2\pi}{T}\right)||^2 \\
     & \text{s.t.} \quad R \in \text{SO}(3), ~ t \in \mathbb{R}^3
\end{split}
\label{eq:panoramic}
\end{align}
where $R_z(\theta)$ denotes the rotation matrix around the $z$-axis by an angle of $\theta$. As in COLMAP~\cite{schoenberger2016sfm}, Levenberg-Marquardt~\cite{triggs2000bundle, hartley2003multiple} is used for optimization. As shown in Figure~\ref{fig:ba_constraint}, this constraint helps correct erroneous camera poses during 3D reconstruction, particularly when working with a limited number of images.

\textbf{Metric scale calibration and ground plane fitting.} Using GPS coordinates of GSV panoramas, we geo-register the \textit{up-to-scale} SfM reconstruction via a 3D similarity transformation optimized between the SfM coordinates and Earth-Centered-Earth-Fixed (ECEF) Cartesian coordinates. This results in a {\it metric scale} 3D scene reconstruction. Subsequently, the road plane is estimated by fitting a plane to the set of 3D points whose 2D pixel locations lie on the \textit{road/lane markings} obtained from an off-the-shelf semantic segmentation method~\cite{cheng2021mask2former}.

\subsection{Camera Localization}
\label{subsection:localization}


Camera localization step aims to determine the intrinsic and extrinsic parameters of the traffic camera with respect to the 3D scene. As depicted in Figure~\ref{fig:calib_pipeline}, we adopt a visual localization pipeline that involves localizing the query image (from the traffic camera) within the 3D reconstruction constructed using GSV images in the previous step.

For every input query image, we retrieve the top-\textit{k} similar database images, where \textit{k} is a predefined value, from the vocabulary tree built in the 3D reconstruction step (Section~\ref{subsection:scene_reconstruction}).
We then match the query image with the $k$ retrieved database images to establish 2D-3D correspondences. For this, we use the learned feature matching method SuperGlue~\cite{sarlin2020superglue} with SuperPoint~\cite{detone2018superpoint} feature descriptors to match the query image with the database images. 
Since our query image is uncalibrated, we follow a sampling-based approach~\cite{irschara2009structure, schoenberger2016sfm} where the pose/focal length is estimated using RANSAC and a minimal pose solver (e.g.,~\cite{gao2003complete, lepetit2009epnp}).
Finally, we perform an extra bundle adjustment step (with panoramic constraints as in Eq.~\ref{eq:panoramic}) to refine both intrinsic parameters and camera poses.

\begin{figure}[ht]
    \centering
    \includegraphics[width=\columnwidth]{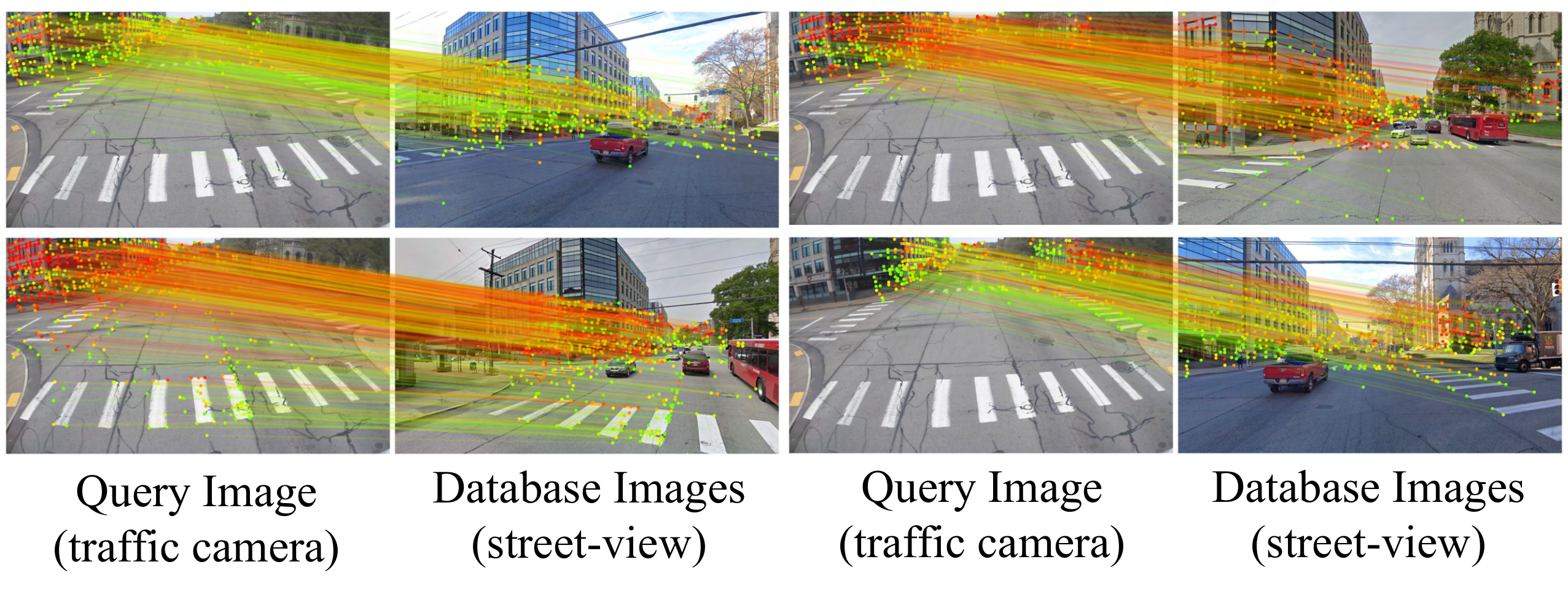}
    \vspace{-5mm}
    \caption{Traffic camera image matched with GSV images using SuperPoint~\cite{detone2018superpoint} and SuperGlue~\cite{sarlin2020superglue}. These methods provide reliable correspondences for accurate absolute pose estimation despite viewpoint and illumination differences.}
    \label{fig:image_matching}
\end{figure}

It is worth noting that the use of learned feature matching is crucial in this step, as it has been shown to outperform hand-crafted feature descriptors and matching methods~\cite{sarlin2020superglue}, particularly in cases where the viewpoint of the traffic camera (often much higher above ground) differs significantly from that of the Google Street View (GSV) images (captured from driving viewpoints). 
Utilizing SuperPoint~\cite{detone2018superpoint} and SuperGlue~\cite{sarlin2020superglue} enables the generation of a large number of accurate matches between the query image and the comprehensive GSV database images (as shown in Figure~\ref{fig:image_matching}). This rich set of matches allows robust recovery of both the intrinsic and extrinsic camera parameters.
Recent advances in local feature matching, such as LightGlue~\cite{lindenberger2023lightglue}, can potentially further improve efficiency.
\section{Experimental Results}
\label{sec:experiments}

\subsection{Calibrated Urban Traffic Cameras (CUTC) Dataset}

\begin{figure}[ht]
    \centering
    \includegraphics[width=0.92\columnwidth]{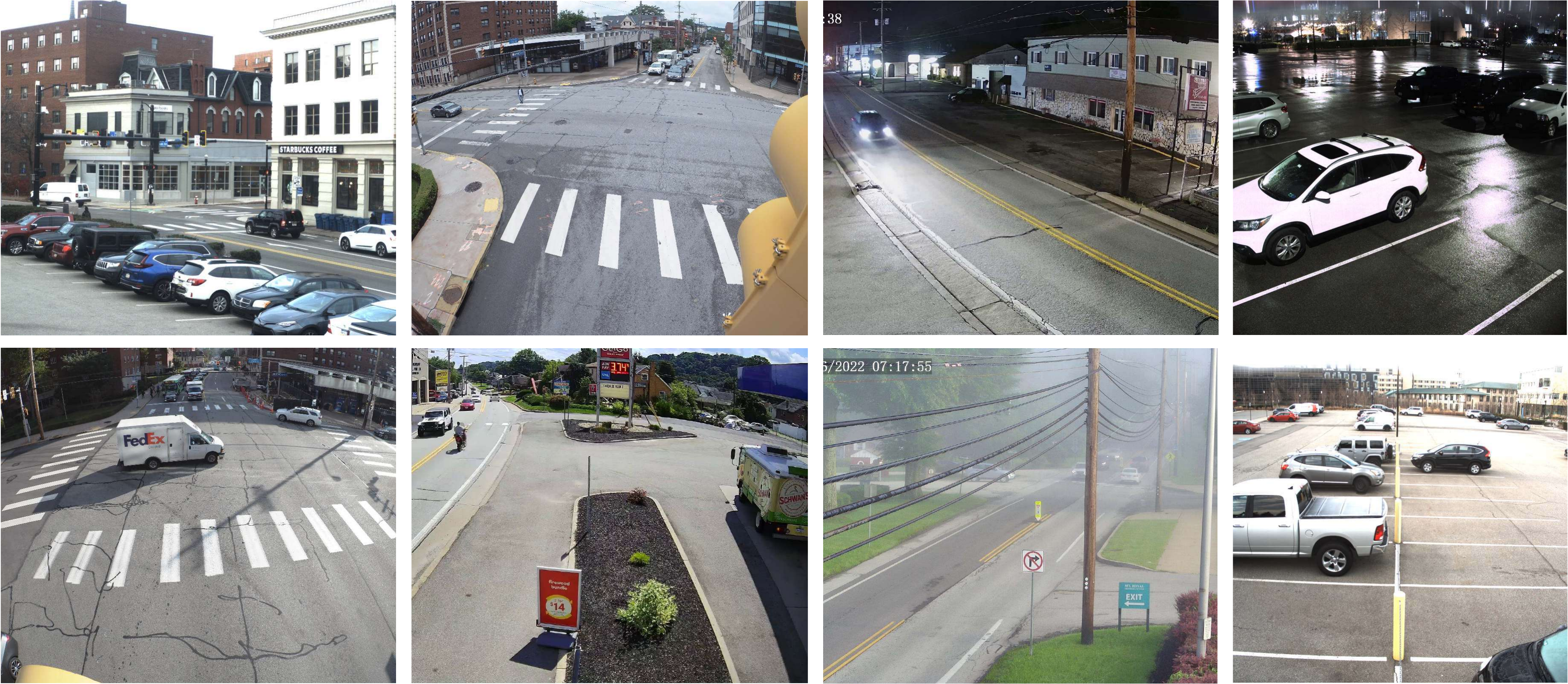}
    \caption{Example images of our Calibrated Urban Traffic Cameras (CUTC) dataset, with diverse  scenes and viewpoints.}
    \label{fig:cucd}
\end{figure}

There are several existing datasets designed for evaluating traffic monitoring algorithms, notably BrnoCompSpeed~\cite{sochor2018comprehensive} and Revaud et al.\cite{revaud2021robust}. However, as mentioned in~\cite{revaud2021robust},  BrnoCompSpeed\cite{sochor2018comprehensive} has limited diversity as the cameras mainly captures vehicles moving in straight lines on highways or freeways. In contrast, Revaud et al.~\cite{revaud2021robust} sought to address this limitation by introducing the CCTV dataset (denoted as Revaud-CCTV dataset), which better emulates real-life CCTV cameras' content and conditions such as low image resolution, non-straight roadways, and imperfect camera lenses. Although both BrnoCompSpeed and Revaud-CCTV provide ground-truth vehicle speeds, the lack of accurate camera calibration restricts its utility for novel traffic analysis applications.

To bridge this gap, we present a new dataset called \textit{Calibrated Urban Traffic Cameras (CUTC)}. This dataset comprises 20 cameras distributed across 6 diverse locations within public urban settings. An example of our CUTC dataset is shown in Figure~\ref{fig:cucd}. Our team installed these cameras which underwent full calibration using checkerboard-based methods~\cite{zhang2000flexible} (\texttt{OpenCV} calibration for \texttt{ChArUco} board) before deployment. Similar to the setup in~\cite{sochor2018comprehensive}, our dataset also incorporates manually measured markers (e.g., lane markings, crosswalks, etc.) on the road plane, with known dimensions between them. These ground-truth measurements serve as reference points for evaluating distance measurements on the ground plane (see Section~\ref{subsec:evalmetrics}).

\subsection{Evaluation Metrics}
\label{subsec:evalmetrics}
\noindent \textbf{Intrinsic Parameters:} Considering the intrinsics parameters computed by checkerboard-based methods~\cite{zhang2000flexible} as ground-truth, we report the mean error (in \%) for focal lengths $(f_x, f_y)$, principal points $(p_x, p_y)$, and distortion coefficients $(k_1, k_2, p_1, p_2)$ over 20 cameras in our CUTC dataset.\\
\noindent \textbf{Ground Distance Measurements:} Following~\cite{bhardwaj2018autocalib, kocur2021traffic}, using manually measured distances between pairs of points on the road plane (e.g., lane markings, crosswalks, etc.) along with their pixel positions in the images, we then computed the normalized error in distance measurement, defined as $r_i = \frac{|\hat{d_i} - d_i|}{\hat{d_i}}$, where $\hat{d_i}$ is the $i$-th ground-truth distance measurement and $d_i$ is the $i$-th measurement based on the ray-plane intersection using the estimated intrinsic matrix and ground-plane equation. This metric effectively gauges the accuracy of both intrinsic and extrinsic parameters.

\subsection{Baseline Methods}
We compare our method to SOTA automatic camera calibration approaches designed specifically for traffic cameras, including OptInOpt~\cite{bartl2019optinopt}, PlaneCalib~\cite{bartl2020planecalib}, DeepVPCalib~\cite{kocur2021traffic}, and Revaud et al.~\cite{revaud2021robust}. In particular, OptInOpt~\cite{bartl2019optinopt} and PlaneCalib~\cite{bartl2020planecalib} rely on localizing 2D landmarks with exact 3D CAD models to infer the focal length of the camera and vehicle poses, DeepVPCalib~\cite{kocur2021traffic} relies on detecting pairs of vanishing points for multiple vehicles in a scene to obtain the focal length of the camera and the orientation of the road plane, and Revaud et al.~\cite{revaud2021robust} learns to predict the calibration (homography between image plane and ground plane) by training solely from synthetic 3D car models.

\subsection{Quantitative Results}

\begin{table}[h]
\centering
\resizebox{\columnwidth}{!}{
\begin{tabular}{l|cccccccc}
\toprule
\textbf{Method}        & $f_x$ & $f_y$ & $p_x$ & $p_y$ & $k_1$ & $k_2$ & $p_1$ & $p_2$ \\ \toprule
OptInOpt~\cite{bartl2019optinopt}      & 11.72     &  11.21    &  2.57*    &  2.61*    & \tikzxmark     & \tikzxmark     & \tikzxmark     &  \tikzxmark    \\
PlaneCalib~\cite{bartl2020planecalib}    & 9.88     &  9.71    &   2.57*   &   2.61*   & \tikzxmark     & \tikzxmark     & \tikzxmark     &  \tikzxmark    \\
DeepVPCalib~\cite{kocur2021traffic}   & 7.51     &   7.33   &   2.57*   &   2.61*   &   \tikzxmark   &  \tikzxmark    & \tikzxmark     &  \tikzxmark    \\
\bottomrule
\textbf{Ours}        &  \textbf{3.17}    &  \textbf{3.54}    &   \textbf{2.11}   & \textbf{2.02}     &  \textbf{7.56}    &  \textbf{8.28}    &  \textbf{6.71}    &  \textbf{8.43 }  
\end{tabular}}
\caption{Comparison between our approach and SOTA techniques in terms of mean error (in \%) of focal lengths $(f_x, f_y)$, principal points $(p_x, p_y)$, and distortion coefficients $(k_1, k_2, p_1, p_2)$. ($\tikzxmark$: unavailable, *: method assumes principal point at image center.)}
\label{table:intrinsics}
\end{table}

\begin{table}[h]
\centering
\resizebox{\columnwidth}{!}{
\begin{tabular}{l|ccc}
\toprule
\textbf{Method}        & Max Error (\%) & Median Error (\%) & RMSE (\%) \\ \toprule
OptInOpt~\cite{bartl2019optinopt}& 15.78           &   10.80           &   12.87   \\
PlaneCalib~\cite{bartl2020planecalib}    &  14.32          &   9.23           &   11.69   \\
DeepVPCalib~\cite{kocur2021traffic}   &  12.17          &       8.11       &   10.62   \\
Revaud et al.~\cite{revaud2021robust} &   14.87         &      10.91        &   12.54   \\
\bottomrule
\textbf{Ours}          &    \textbf{6.75}        &     \textbf{3.22}         &     \textbf{4.68}
\end{tabular}}
\caption{Comparison between our approach and SOTA automatic calibration techniques in terms of max, median, and RMSE (in \%) between measured vs. estimated distances on ground plane.}
\label{table:extrinsics}
\end{table}


\noindent \textbf{Intrinsics Parameters:} In Table~\ref{table:intrinsics}, using checkerboard-based calibration as ground-truth, we compare our approach against SOTA automatic calibration techniques~\cite{bartl2019optinopt, bartl2020planecalib, kocur2021traffic}. Our method significantly outperforms SOTA methods due to the fact that the dense coverage of GSV images (with \textit{known} intrinsics parameters) and reliable correspondences from learned feature matching allows us to register the traffic camera into the 3D scene with high accuracy. Conversely, methods like OptInOpt~\cite{bartl2019optinopt} and PlaneCalib~\cite{bartl2020planecalib}, relying on precise 3D CAD models of vehicles, exhibit diminished generalization accuracy when faced with cameras in diverse countries with unknown vehicle type. Additionally, we observe that the focal length estimation from DeepVPCalib~\cite{kocur2021traffic} is instable as it is highly sensitive to the accuracy of detected vanishing points. Lastly, these  methods assume no distortion, an assumption that may not hold in practice. Hence, an additional strength of our approach is its capability to estimate distortion parameters (radial and tangential) with a satisfactory accuracy level (within $10\%$).

\noindent\textbf{Distance Measurements:} Using manually measured distances on the ground plane, in Table~\ref{table:extrinsics}, we report the max, median, and root-mean-squared error (in \%) over all of the possible pairs of ground truth measurements in our CUTC dataset. For methods that do not directly infer metric scene scale such as DeepVPCalib~\cite{kocur2021traffic}, we scale the estimated distances with the ground-truth scale. As shown in Table~\ref{table:extrinsics}, our method outperforms existing SOTA methods by a large margin, demonstrating the accuracy of our camera calibration as well as estimated scene geometry. In our experiments, we observe limited generalizability from the pretrained model of Revaud et al.~\cite{revaud2021robust} that was trained exclusively on synthetic 3D car models.


\begin{figure}
    \centering
        \includegraphics[width=0.9\columnwidth]{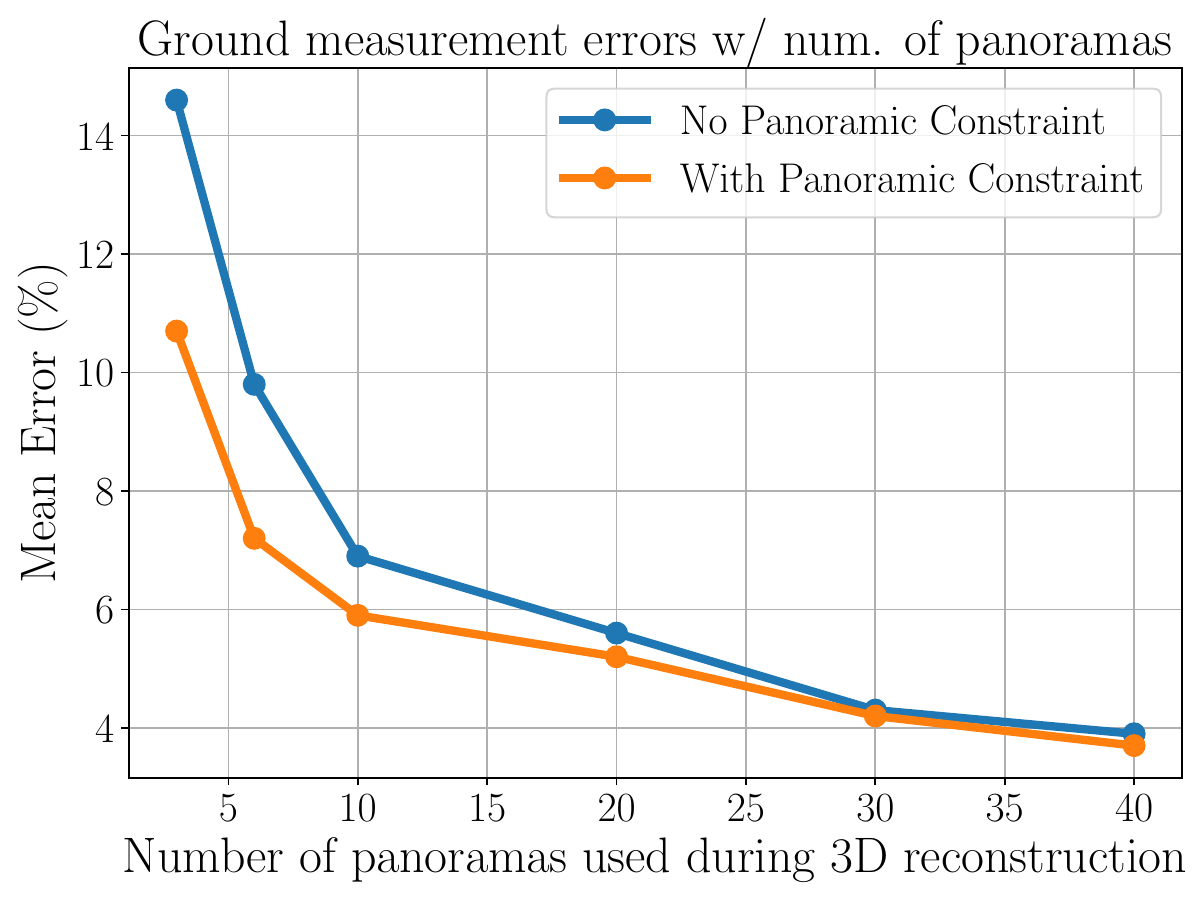}
    \vspace{-2mm}
    \caption{We show the mean errors on estimating ground distances w.r.t. the number of panoramas being used during 3D reconstruction. By enforcing panoramic constraint during reconstruction, our method improves the accuracy significantly over the baseline, especially in small number of views.}
    \label{fig:pano_constraint}
\end{figure}

\begin{figure*}[ht]
    \centering
    \includegraphics[width=0.95\textwidth]{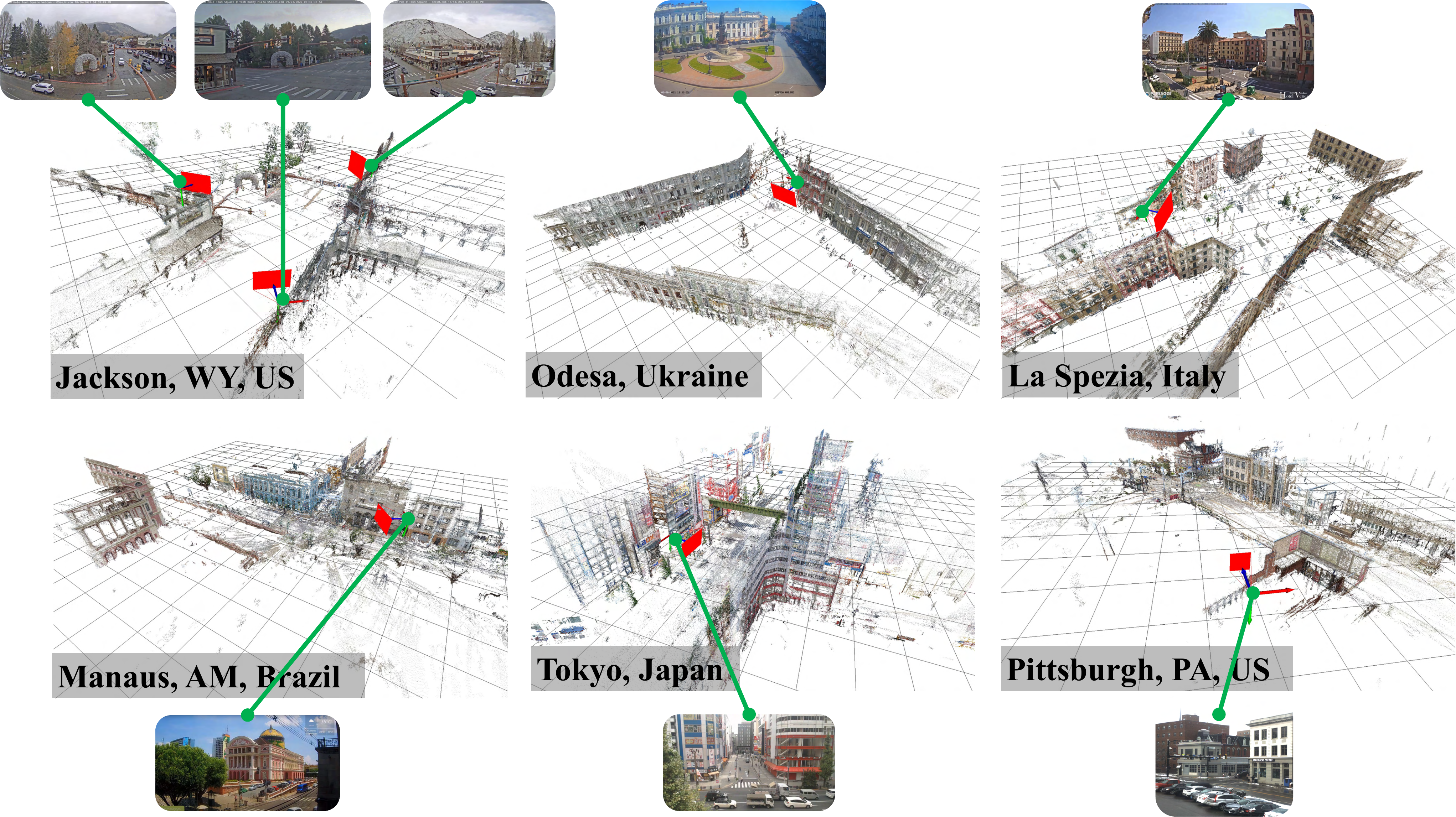}
    \vspace{-2mm}
    \caption{Additional examples demonstrate our method's robustness in reconstructing scenes and localizing cameras spanning various countries and continents (traffic camera visualized in \textcolor{red}{red}).}
    \label{fig:map_loc}
\end{figure*}

\noindent \textbf{Enforcing the known relative pose between frames from the same panorama improves the accuracy of reconstruction}: While it is well-known that using more images during 3D reconstruction leads to better camera calibration accuracy, our key observation is that by enforcing the known relative pose between frames from the same panorama during 3D reconstruction, we can significantly boost accuracy, especially when the number of GSV panoramas being used for reconstruction is limited (as shown in Figure~\ref{fig:pano_constraint}). This becomes crucial when dealing with the limited availability of GSV images at a location, or when computational efficiency is a priority.

\subsection{Qualitative Results} Our camera localization method proves versatile across diverse cameras in real-world settings. As depicted in Figure~\ref{fig:map_loc}, we successfully achieve both accurate 3D scene reconstruction and precise camera localization across different locations spanning multiple countries and continents.
Importantly, our framework's adaptability extends beyond Google Street View, making it highly versatile for different street-level imagery sources~\cite{BingStreetside, Mapillary}, with the potential to achieve camera calibration on a global scale.


\begin{figure*}[ht]
\centering
{\includegraphics[width=0.92\textwidth]{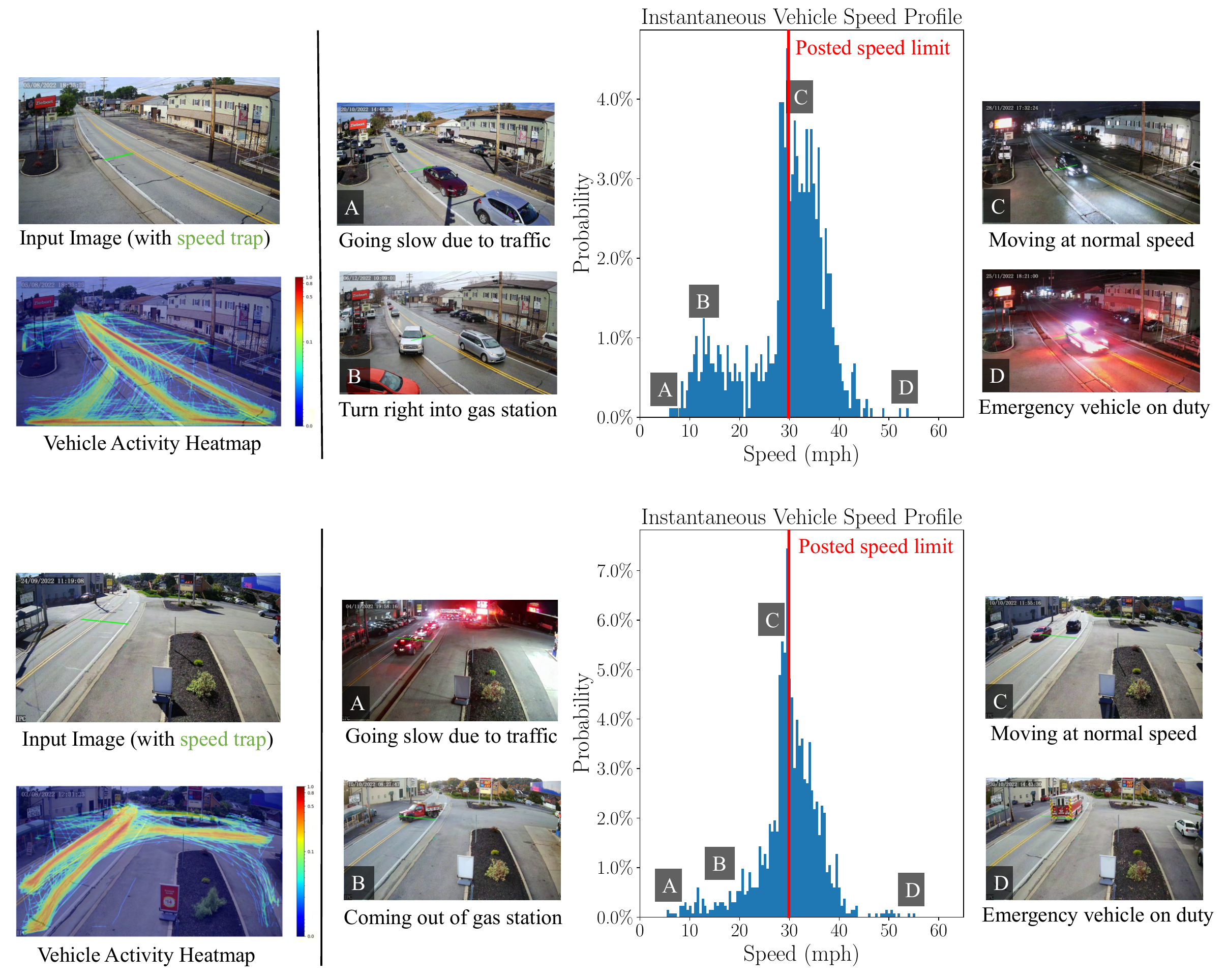}}
\vspace{-3mm}
\caption{Speed estimates and activity heatmaps depicted for two distinct virtual speed traps (indicated by \textcolor{green}{green lines}) across two different cameras. Various scenarios are displayed, including slow traffic due to congestion and fast movement of emergency vehicles.}
\label{fig:apps_speed1}
\end{figure*}

\section{Applications}
\label{sec:applications}


With accurate camera calibration information, we show its applications in 3D reconstruction and speed measurements of moving vehicles, allowing us to gain unique insights that can inform decision-making processes related to traffic management and safety measures.

In Figure~\ref{fig:apps_speed1}, we provide automatic vehicle speed estimates and activity heatmaps for two different cameras. First, we use an off-the-shelf object detector~\cite{MaskRCNN} and tracker~\cite{Bewley2016_sort} to compute the tracked detections of every vehicle.
Using vehicles' active mean shape models and detected 2D keypoints~\cite{onet_cvpr19, ke2020gsnet}, following~\cite{li2021traffic4d}, we optimize for the 6DoF vehicle pose and shape variations (defined by PCA components of the mean shape model) for each vehicle track by enforcing all the detected objects to lie on the ground plane.

\noindent \textbf{Activity Heatmap:} Heatmaps visualize the level of vehicle activity at each camera location. These heatmaps are generated by aggregating the centroid of the 3D tracks of all vehicles over the entire data acquisition period, then normalized by the maximum count, resulting in a value ranging from 0 (blue, no activity) to 1 (red, high activity).

\noindent \textbf{Vehicle Speed:}
For each camera, we created virtual speed traps (visualized as a \textcolor{green}{green line} on the road) that allows estimates of speed a vehicle crossing over the region of interest.
Consequently, the reported speeds correspond to instantaneous speed readings garnered from the virtual speed trap.
By leveraging the reconstructed 3D shape and pose of vehicles, we calculate the velocity as the front of the vehicle crosses the virtual speed trap on the ground plane.
As depicted in Figure~\ref{fig:apps_speed1}, precise camera calibration facilitates the accurate measurement of vehicle speeds, enabling us to automatically derive valuable insights from the data. For instance, the system can identify prevalent traffic patterns to enhance urban planning or detect anomalies such as accidents or high-speed emergency vehicles on duty.
Importantly, the benefits of accurate metric 3D scene reconstruction and camera calibration extend beyond speed estimation as they are crucial for various applications, including  understanding human-vehicle interaction for accident prediction and prevention~\cite{loewenherz2017video}, achieving multi-camera fusion by aligning different cameras' views to a common frame~\cite{1413272, bhardwaj2018autocalib}, and so on.



\section{Discussion}
\label{sec:conclusions}

We have presented a scalable framework that leverages street-level imagery for metric 3D model reconstruction, enabling accurate calibration of real-world traffic cameras. Our approach can be applied to any camera with sufficient nearby street-level imagery, making it practical to be used worldwide. We show the framework's value in traffic analysis through insights derived from 3D vehicle reconstruction and speed measurement, providing valuable information for improving transportation systems and urban infrastructure.

\noindent \textbf{Potential Societal Impact:} We do not perform any human subject studies from these cameras. To preserve the privacy of the object captured in the images, we blur the faces and license plates in all the images to be released. This study is designated as non-human subjects research by our Institutional Review Board (IRB).

\noindent \textbf{Limitations:} Our method requires capturing at least some portion of the scene's ``background'' for feature matching. Thus, scenes with severely limited contextual information, e.g., situations where cameras are oriented to solely capture freeway surfaces while looking straight down, can hinder the performance of our approach.

\small \noindent \textbf{Acknowledgements:}  This work was supported in part by an NSF Grant CNS-2038612, a DOT RITA Mobility-21 Grant 69A3551747111, and Intelligence Advanced Research Projects Activity (IARPA) via Department of Interior/ Interior Business Center (DOI/IBC) contract number 140D0423C0074. The U.S. Government is authorized to reproduce and distribute reprints for Governmental purposes notwithstanding any copyright annotation thereon. Disclaimer: The views and conclusions contained herein are those of the authors and should not be interpreted as necessarily representing the official policies or endorsements, either expressed or implied, of IARPA, DOI/IBC, or the U.S. Government.

{\small
\bibliographystyle{ieee_fullname}
\bibliography{egbib}

\begin{thebibliography}{10}\itemsep=-1pt

\bibitem{arandjelovic2013all}
Relja Arandjelovic and Andrew Zisserman.
\newblock All about vlad.
\newblock In {\em CVPR}, 2013.

\bibitem{10.5555/1283383.1283494}
David Arthur and Sergei Vassilvitskii.
\newblock K-means++: The advantages of careful seeding.
\newblock In {\em SODA}, 2007.

\bibitem{bartl2019optinopt}
Vojt{\v{e}}ch Bartl and Adam Herout.
\newblock Optinopt: Dual optimization for automatic camera calibration by
  multi-target observations.
\newblock In {\em AVSS}, 2019.

\bibitem{bartl2020planecalib}
Vojt{\v{e}}ch Bartl, Roman Juranek, Jakub {\v{S}}pa{\v{n}}hel, and Adam Herout.
\newblock Planecalib: Automatic camera calibration by multiple observations of
  rigid objects on plane.
\newblock In {\em DICTA}, 2020.

\bibitem{Bewley2016_sort}
Alex Bewley, Zongyuan Ge, Lionel Ott, Fabio Ramos, and Ben Upcroft.
\newblock Simple online and realtime tracking.
\newblock In {\em ICIP}, 2016.

\bibitem{bhardwaj2018autocalib}
Romil Bhardwaj, Gopi~Krishna Tummala, Ganesan Ramalingam, Ramachandran Ramjee,
  and Prasun Sinha.
\newblock Autocalib: Automatic traffic camera calibration at scale.
\newblock {\em TOSN}, 2018.

\bibitem{bi2019joint}
Huikun Bi, Zhong Fang, Tianlu Mao, Zhaoqi Wang, and Zhigang Deng.
\newblock Joint prediction for kinematic trajectories in
  vehicle-pedestrian-mixed scenes.
\newblock In {\em CVPR}, 2019.

\bibitem{BingStreetside}
Bing.
\newblock {Bing Streetside}.
\newblock https://www.bing.com/maps/.

\bibitem{bogdan2018deepcalib}
Oleksandr Bogdan, Viktor Eckstein, Francois Rameau, and Jean-Charles Bazin.
\newblock Deepcalib: A deep learning approach for automatic intrinsic
  calibration of wide field-of-view cameras.
\newblock In {\em Proceedings of the 15th ACM SIGGRAPH European Conference on
  Visual Media Production}, pages 1--10, 2018.

\bibitem{cathey2005}
Joseph~R Cathey and Matthew~A Dailey.
\newblock Camera calibration using lane markings: An evaluation of vanishing
  point detection methods.
\newblock {\em IEEE Transactions on Intelligent Transportation Systems},
  6(2):124--133, 2005.

\bibitem{cheng2021mask2former}
Bowen Cheng, Ishan Misra, Alexander~G. Schwing, Alexander Kirillov, and Rohit
  Girdhar.
\newblock Masked-attention mask transformer for universal image segmentation.
\newblock 2022.

\bibitem{dailey2000}
Matthew~A Dailey, Benjamin~C Schoepflin, Juraj Sochor, and Michal Seman.
\newblock Camera calibration for traffic scene analysis using vehicle motion.
\newblock {\em IEEE Transactions on Intelligent Transportation Systems},
  1(1):43--50, 2000.

\bibitem{detone2018superpoint}
Daniel DeTone, Tomasz Malisiewicz, and Andrew Rabinovich.
\newblock Superpoint: Self-supervised interest point detection and description.
\newblock In {\em Proceedings of the IEEE conference on computer vision and
  pattern recognition workshops}, pages 224--236, 2018.

\bibitem{do2015}
Hoang~Dung Do and Reinhard Klette.
\newblock Camera calibration for road scene analysis using vehicle motion and
  lane markings.
\newblock {\em IEEE Transactions on Intelligent Transportation Systems},
  16(7):2700--2712, 2015.

\bibitem{dubska2014}
Katerina Dubska, Jiri Matas, Ondrej Holik, and Michal Seman.
\newblock Camera calibration using vehicle motion.
\newblock {\em IEEE Transactions on Intelligent Transportation Systems},
  15(1):283--294, 2014.

\bibitem{dubska2015}
Katerina Dubska, Jiri Matas, Ondrej Holik, and Michal Seman.
\newblock Camera calibration for road scene analysis using vehicle motion with
  robust estimation of camera parameters.
\newblock {\em IEEE Transactions on Intelligent Transportation Systems},
  16(2):540--551, 2015.

\bibitem{friedman1977algorithm}
Jerome~H Friedman, Jon~Louis Bentley, and Raphael~Ari Finkel.
\newblock An algorithm for finding best matches in logarithmic expected time.
\newblock {\em ACM Transactions on Mathematical Software (TOMS)},
  3(3):209--226, 1977.

\bibitem{gao2003complete}
Xiao-Shan Gao, Xiao-Rong Hou, Jianliang Tang, and Hang-Fei Cheng.
\newblock Complete solution classification for the perspective-three-point
  problem.
\newblock {\em IEEE transactions on pattern analysis and machine intelligence},
  25(8):930--943, 2003.

\bibitem{giannakeris2018speed}
Panagiotis Giannakeris, Vagia Kaltsa, Konstantinos Avgerinakis, Alexia
  Briassouli, Stefanos Vrochidis, and Ioannis Kompatsiaris.
\newblock Speed estimation and abnormality detection from surveillance cameras.
\newblock In {\em Proceedings of the IEEE Conference on Computer Vision and
  Pattern Recognition Workshops}, pages 93--99, 2018.

\bibitem{GSVAnniversary}
Google.
\newblock {Celebrating 15 years of Street View}.
\newblock https://www.google.com/streetview/anniversary/.

\bibitem{GoogleStreetView}
Google.
\newblock {Google Street View}.
\newblock https://www.google.com/streetview/.

\bibitem{grammatikopoulos2005}
Vasileios Grammatikopoulos, Manolis~N Lourakis, and John~K Tsotsos.
\newblock Camera calibration for road scene analysis using vanishing points.
\newblock {\em IEEE Transactions on Intelligent Transportation Systems},
  6(2):134--144, 2005.

\bibitem{hartley2003multiple}
Richard Hartley and Andrew Zisserman.
\newblock {\em Multiple view geometry in computer vision}.
\newblock Cambridge university press, 2003.

\bibitem{he2007b}
Jiebo He and Nicholas Yung.
\newblock Camera calibration for traffic scene analysis using lane markings.
\newblock {\em IEEE Transactions on Intelligent Transportation Systems},
  8(3):417--427, 2007.

\bibitem{MaskRCNN}
Kaiming He, Georgia Gkioxari, Piotr Doll{\'a}r, and Ross Girshick.
\newblock Mask r-cnn.
\newblock In {\em ICCV}, 2017.

\bibitem{irschara2009structure}
Arnold Irschara, Christopher Zach, Jan-Michael Frahm, and Horst Bischof.
\newblock From structure-from-motion point clouds to fast location recognition.
\newblock In {\em 2009 IEEE Conference on Computer Vision and Pattern
  Recognition}, pages 2599--2606. IEEE, 2009.

\bibitem{ke2020gsnet}
Lei Ke, Shichao Li, Yanan Sun, Yu-Wing Tai, and Chi-Keung Tang.
\newblock Gsnet: Joint vehicle pose and shape reconstruction with geometrical
  and scene-aware supervision.
\newblock In {\em European Conference on Computer Vision}, pages 515--532.
  Springer, 2020.

\bibitem{kocur2021traffic}
Viktor Kocur and Milan Ft{\'a}{\v{c}}nik.
\newblock Traffic camera calibration via vehicle vanishing point detection.
\newblock In {\em ICANN}, pages 628--639. Springer, 2021.

\bibitem{lan2014}
Man Lan, Jiang Zhao, and Tiantian Zhu.
\newblock Camera calibration for traffic scene analysis using multiple
  vanishing points.
\newblock {\em IEEE Transactions on Intelligent Transportation Systems},
  15(4):1806--1817, 2014.

\bibitem{lepetit2009epnp}
Vincent Lepetit, Francesc Moreno-Noguer, and Pascal Fua.
\newblock Epnp: An accurate o (n) solution to the pnp problem.
\newblock {\em International journal of computer vision}, 81(2):155, 2009.

\bibitem{li2021traffic4d}
Fangyu Li, N~Dinesh Reddy, Xudong Chen, and Srinivasa~G Narasimhan.
\newblock Traffic4d: Single view longitudinal 4d reconstruction of repetitious
  activity using self-supervised experts.
\newblock In {\em IEEE Intelligent Vehicles Symposium}, 2021.

\bibitem{lindenberger2023lightglue}
Philipp Lindenberger, Paul-Edouard Sarlin, and Marc Pollefeys.
\newblock {LightGlue: Local Feature Matching at Light Speed}.
\newblock In {\em ICCV}, 2023.

\bibitem{loewenherz2017video}
Franz Loewenherz, Victor Bahl, and Yinhai Wang.
\newblock Video analytics towards vision zero.
\newblock {\em Institute of Transportation Engineers. ITE Journal}, 87(3):25,
  2017.

\bibitem{lopez2019deep}
Manuel Lopez, Roger Mari, Pau Gargallo, Yubin Kuang, Javier Gonzalez-Jimenez,
  and Gloria Haro.
\newblock Deep single image camera calibration with radial distortion.
\newblock In {\em Proceedings of the IEEE/CVF Conference on Computer Vision and
  Pattern Recognition}, pages 11817--11825, 2019.

\bibitem{luvizon2014}
Luiz~Henrique Luvizon, Marcelo de Souza, and Mohamed Bennamoun.
\newblock Camera calibration for traffic scene analysis using vehicle motion.
\newblock {\em IEEE Transactions on Intelligent Transportation Systems},
  15(1):363--374, 2014.

\bibitem{luvizon2016}
Luiz~Henrique Luvizon, Marcelo de Souza, and Mohamed Bennamoun.
\newblock Camera calibration for traffic scene analysis using vehicle motion
  with uncertainty estimation.
\newblock {\em IEEE Transactions on Intelligent Transportation Systems},
  17(1):270--281, 2016.

\bibitem{maduro2008}
Ricardo Maduro and Reinhard Klette.
\newblock Camera calibration for road scene analysis using manual measurements.
\newblock {\em IEEE Transactions on Intelligent Transportation Systems},
  9(3):501--510, 2008.

\bibitem{Mapillary}
Mapillary.
\newblock {Mapillary Maps}.
\newblock https://www.mapillary.com/.

\bibitem{1413272}
K. Muller, A. Smolic, M. Drose, P. Voigt, and T. Wiegand.
\newblock 3-d reconstruction of a dynamic environment with a fully calibrated
  background for traffic scenes.
\newblock {\em IEEE Transactions on Circuits and Systems for Video Technology},
  15(4):538--549, 2005.

\bibitem{nubert2018traffic}
Julian Nubert, Nicholas~Giai Truong, Abel Lim, Herbert~Ilhan Tanujaya, Leah
  Lim, and Mai~Anh Vu.
\newblock Traffic density estimation using a convolutional neural network.
\newblock {\em arXiv preprint arXiv:1809.01564}, 2018.

\bibitem{nurhadiyatna2013}
Angga Nurhadiyatna and Reinhard Klette.
\newblock Camera calibration for road scene analysis using manual measurements.
\newblock {\em IEEE Transactions on Intelligent Transportation Systems},
  14(3):1149--1159, 2013.

\bibitem{onet_cvpr19}
N.~Dinesh Reddy, Minh Vo, and Srinivasa~G. Narasimhan.
\newblock Occlusion-net: 2d/3d occluded keypoint localization using graph
  networks.
\newblock In {\em The IEEE Conference on Computer Vision and Pattern
  Recognition (CVPR)}, pages 7326--7335, 2019.

\bibitem{revaud2021robust}
Jerome Revaud and Martin Humenberger.
\newblock Robust automatic monocular vehicle speed estimation for traffic
  surveillance.
\newblock In {\em Proceedings of the IEEE/CVF International Conference on
  Computer Vision}, pages 4551--4561, 2021.

\bibitem{sabbani2018deep}
Imad Sabbani, Andres Perez-Uribe, Omar Bouattane, and Abdellah El~Moudni.
\newblock Deep convolutional neural network architecture for urban traffic flow
  estimation.
\newblock {\em IJCSNS International Journal of Computer Science and Network
  Security}, 2018.

\bibitem{sarlin2020superglue}
Paul-Edouard Sarlin, Daniel DeTone, Tomasz Malisiewicz, and Andrew Rabinovich.
\newblock Superglue: Learning feature matching with graph neural networks.
\newblock In {\em Proceedings of the IEEE/CVF Conference on Computer Vision and
  Pattern Recognition}, pages 4938--4947, 2020.

\bibitem{schoepflin2003}
Benjamin~C Schoepflin and Matthew~A Dailey.
\newblock Camera calibration for traffic scene analysis using vehicle motion.
\newblock {\em IEEE Transactions on Intelligent Transportation Systems},
  4(2):111--120, 2003.

\bibitem{schoenberger2016sfm}
Johannes~Lutz Sch\"{o}nberger and Jan-Michael Frahm.
\newblock Structure-from-motion revisited.
\newblock In {\em Conference on Computer Vision and Pattern Recognition
  (CVPR)}, 2016.

\bibitem{shah2018cadp}
Ankit~Parag Shah, Jean-Bapstite Lamare, Tuan Nguyen-Anh, and Alexander
  Hauptmann.
\newblock Cadp: A novel dataset for cctv traffic camera based accident
  analysis.
\newblock In {\em 2018 15th IEEE International Conference on Advanced Video and
  Signal Based Surveillance (AVSS)}, pages 1--9. IEEE, 2018.

\bibitem{sina2013}
Mostafa Sina and Reinhard Klette.
\newblock Camera calibration for road scene analysis using manual measurements.
\newblock {\em IEEE Transactions on Intelligent Transportation Systems},
  14(4):1595--1604, 2013.

\bibitem{sochor2017traffic}
Jakub Sochor, Roman Jur{\'a}nek, and Adam Herout.
\newblock Traffic surveillance camera calibration by 3d model bounding box
  alignment for accurate vehicle speed measurement.
\newblock {\em Computer Vision and Image Understanding}, 161:87--98, 2017.

\bibitem{sochor2018comprehensive}
Jakub Sochor, Roman Jur{\'a}nek, Jakub {\v{S}}pa{\v{n}}hel, Luk{\'a}{\v{s}}
  Mar{\v{s}}{\'\i}k, Adam {\v{S}}irok{\`y}, Adam Herout, and Pavel
  Zem{\v{c}}{\'\i}k.
\newblock Comprehensive data set for automatic single camera visual speed
  measurement.
\newblock {\em IEEE Transactions on Intelligent Transportation Systems},
  20(5):1633--1643, 2018.

\bibitem{triggs2000bundle}
Bill Triggs, Philip~F McLauchlan, Richard~I Hartley, and Andrew~W Fitzgibbon.
\newblock Bundle adjustment—a modern synthesis.
\newblock In {\em Vision Algorithms: Theory and Practice: International
  Workshop on Vision Algorithms Corfu, Greece, September 21--22, 1999
  Proceedings}, pages 298--372. Springer, 2000.

\bibitem{tsai1987versatile}
Roger Tsai.
\newblock A versatile camera calibration technique for high-accuracy 3d machine
  vision metrology using off-the-shelf tv cameras and lenses.
\newblock {\em IEEE Journal on Robotics and Automation}, 3(4):323--344, 1987.

\bibitem{wang2007research}
Kunfeng Wang, Hua Huang, Yuantao Li, and Fei-Yue Wang.
\newblock Research on lane-marking line based camera calibration.
\newblock In {\em 2007 IEEE International Conference on Vehicular Electronics
  and Safety}, pages 1--6. IEEE, 2007.

\bibitem{zhang2022monocular}
Xingchen Zhang, Yuxiang Feng, Panagiotis Angeloudis, and Yiannis Demiris.
\newblock Monocular visual traffic surveillance: A review.
\newblock {\em IEEE Transactions on Intelligent Transportation Systems},
  23(9):14148--14165, 2022.

\bibitem{zhang2000flexible}
Zhengyou Zhang.
\newblock A flexible new technique for camera calibration.
\newblock {\em IEEE Transactions on pattern analysis and machine intelligence},
  22(11):1330--1334, 2000.

\end{thebibliography}
}

\end{document}